%% file: eacl2017.tex
\documentclass[11pt]{article}
\usepackage{eacl2017}
\usepackage{times}
\usepackage{url}
\usepackage{latexsym}
\usepackage{verbatim}
\usepackage{tikz-qtree, algorithm, algorithmic, adjustbox}

\eaclfinalcopy % Uncomment this line for the final submission
\title{Grammatical Constraints on Intra-sentential Code-Switching:\\  From Theories to Working Models}

\author{Gayatri H. Bhat\thanks{This work was done when the author was a Research Intern at Microsoft Research Lab India.}\\
  Birla Institute of Technology \\
  and Science, Pilani\\
  {\tt f2013087@pilani.bits-}\\
  {\tt pilani.ac.in} \\\And
  Monojit Choudhury\\
  Microsoft Research Labs,\\
  Bangalore, India \\
  {\tt monojitc}\\
  {\tt @microsoft.com} \\\And
  Kalika Bali\\
  Microsoft Research Labs,\\
  Bangalore, India \\
  {\tt kalikab}\\
  {\tt @microsoft.com}
}

\date{}

\begin{document}
\maketitle
\begin{abstract}
 We make one of the first attempts to build working models for intra-sentential code-switching based on the {\em Equivalence-Constraint}~\cite{poplack-1980} and {\em Matrix-Language}~\cite{Myers-Scotton1993} theories. We conduct a detailed theoretical analysis, and a small-scale empirical study of the two models for Hindi-English CS. Our analyses show that the models are neither sound nor complete. Taking insights from the errors made by the models, we propose a new model that combines features of both the theories.
\end{abstract}

\input{1.introduction.tex}

\input{2.theoretical_models.tex}

\input{3.implementation.tex}

\input{4.emperical_analysis.tex}

\input{5._discussion_litsurvey.tex}

\bibliography{code_switch}
\bibliographystyle{eacl2017}

\appendix
\input{glosses.tex}
\input{algorithms.tex}

\end{document}

%% file: 1.introduction.tex
\section{Introduction}
{\em Code-Switching} (CS) is defined as the juxtaposition of words and fragments from two or more languages in a single conversation or utterance~\cite{poplack-1980}. Linguistic studies on intra-sentential CS have indicated beyond doubt that there are lexical and grammatical constraints on switching, though there is much debate and many schools of thought on what these constraints are (see Muysken~\cite{Muysken1995} and references there in). In recent times, computational processing of CS has received much attention~\cite{solorio2008part,adel2013combination,vyas-emnlp-2014,solorio:2014,elfardy2014aida,adel2015syntactic,sharma2016shallow}, primarily for two reasons. First, user-generated content on social media from multilingual communities is often code-switched~\cite{das2013code,bali-2014}, and second, speech based interfaces such as conversational agents for multilingual societies also need to handle CS in speech~\cite{huang2014improving,yilmaz2016investigating}. However, despite the longstanding linguistic research in the area, we do not know of any study that attempts to build computational models of the grammatical constraints on intra-sentential CS.

Such models could be useful for (1) automatic generation of grammatically well-formed and natural CS text, which in turn can help in training large scale language models for CS (the alternative approach of learning language models directly from data is severely limited due to absence of large-scale CS corpora), (2) validation of or comparison between the various linguistic theories of CS, which could potentially include large-scale data-driven analysis, and (3) parsing of CS text leading to a better understanding and processing of the data.

In this paper, we build computational models for two of the most popular linguistic theories of intra-sentential CS, namely the {\em Equivalence-Constraint} (EC) theory~\cite{poplack-1980,sankoff1998} and the {\em Matrix-Language} (ML) theory~\cite{Joshi85,Myers-Scotton1993,Myers-Scotton1995}. While these theories are well-defined, several important aspects that are necessary for the  implementation of a working model remain unspecified. We make a series of systematic assumptions that help us build configurable and language-independent working models. We then conduct a detailed theoretical analysis of the models and a small-scale empirical experiment on the acceptability of the generated sentences for Hindi-English CS. Our analyses agree with the existing linguistic literature which argues that both theories are neither sound nor complete, though they certainly provide useful insights into lexical and grammatical constraints on CS. Finally, we propose a more effectual computational model of CS that combines certain constraints from the EC model with a relaxed version of the ML model.      

The rest of the paper is organized as follows. Sec. 2 formally defines the EC and ML models and introduces the notions of theoretical and empircial equivalence of such models that provide us with a framework to analyze and compare the models. Sec. 3 discusses the implementation of the two models. In Sec. 4 we present an empirical study. We conclude in Sec. 5 by discussing various interesting issues and open problems in this area.

%% file: 2.theoretical_models.tex
\section{Formal Description of the Models}

The grammatical theories of CS can be broadly classified into {\em alternational} and {\em insertional} approaches~\cite{Muysken1995}. The {\em Equivalence-Constraint} (EC) theory~\cite{poplack-1980,sankoff1998} , one of the first and a popular  alternational approach, holds that in a well-formed CS sentence, each monolingual fragment should be well-formed with respect to its own grammar, and switching is allowed only at those points where the grammatical constraints of both the languages are satisfied. The {\em Matrix-Language} (ML) theory~\cite{Joshi85,Myers-Scotton1993,Myers-Scotton1995}, on the other hand, is an insertional account of CS,  according to which the structure of any mixed language sentence is governed by the grammar of a single language, called the {\em matrix language}. One or more constituents of another language (aka the {\em embedded language}) can be inserted or embedded within the matrix.

Several decades of research have challenged these models by presenting counter-examples from various language pairs. This led to modifications of the theories or alternative proposals, e.g., the government based accounts of CS~\cite{Discuillo1986}. The question is far from settled and the status of research today is hardly any different from what Pieter Muysken~\shortcite{Muysken1995} had summarized two decades ago: ``{\it the present state of the field [is] characterised by pluralism and the growing recognition that various mechanisms may play a role in different code-switching situations.}''

Among the theoretical treatments of intra-sentential CS, the ones by Joshi~\shortcite{Joshi85} on the ML framework and Sankoff~\shortcite{sankoff1998} on the theory of EC are defined at a level of granularity and style that are most readily adaptable for implementation. Hence, we choose to implement and analyze these models, and refer to them generically as the ML and EC models, though the reader should keep in mind that the models have been subsequently modified~\cite{Myers-Scotton1995} and our treatment here does not consider the most recent versions of the theories.  

\subsection{Notations and Assumptions}
Consider two languages $L_1$ and $L_2$, defined by context-free grammars $G_1$=$\langle V_1, \Sigma_1, R_1, S_1 \rangle$ and $G_2$=$\langle V_2, \Sigma_2, R_2, S_2 \rangle$ respectively. An intra-sentential CS sentence in $L_1$ and $L_2$ is not well-formed according to either $G_1$ or $G_2$, and therefore is a part of neither $L_1$ nor $L_2$. Let us denote the set of such sentences, the $L_1$-$L_2$ CS language, as $L_X$. The corresponding hybrid grammar $G_X$ is some composition of $G_1$ and $G_2$, say $G_1\otimes G_2$. The fundamental question in the grammatical theory of CS is ``what is this composition $\otimes$?".

In order to answer this question, both ML and EC models make a common assumption of  

{\bf Categorical congruence}, according to which there exists a mapping $f_{12}$ such that every $v \in V_1$ has a corresponding $f_{12}(v) \in V_2$ (and another mapping $f_{21}$ defined vice-versa). The EC theory further requires these mappings to be bijections, with $f_{12}^{-1} = f_{21}$. We shall refer to this as {\em Strong Categorical Congruence} (SCC). 

The EC model further assumes that
\textit{every code-mixed sentence in $L_X$ is a composition of a pair of semantically and syntactically equivalent monolingual sentences}. This, in fact, follows directly from SCC, and does not require any further assumption apart from some loose lexical congruence. 
At this point, it might be useful to look at a real example. {\bf 1H} and {\bf 1E} show a pair of semantically and syntactically equivalent sentences in Hindi (Hi) and English (En) respectively, and {\bf 1a} shows an acceptable code-switched Hi-En sentence.\footnote{The following conventions are followed here: {\bf H} and {\bf E} represent the (equivalent) monolingual  Hi and En sentences. Lower case letters, {\bf a}, {\bf b}, etc., represent code-switched sentences. Hindi words are Romanized and shown in italics.}

\vspace*{-0.5\baselineskip}
\begin{table}[H]
\begin{tabular}{lcccc}
\bf 1H. & \tt\itshape Shanivar & \tt{\itshape neeras} & \tt\itshape hai & \tt\itshape uss \\ 
& Saturday & boring & is & that \\ 
 & \tt\itshape nazariye & \tt\itshape se. \\
& perspective & from
\end{tabular}
\end{table}
\vspace*{-\baselineskip} \par
 
\vspace*{-0.5\baselineskip} 
\begin{table}[H]
\begin{tabular}{ll}
\bf 1E. & \tt Saturday is boring from that \\
& \tt perspective.
\end{tabular}
\end{table}
\vspace*{-\baselineskip} \par 

\vspace*{-0.5\baselineskip} 
\begin{table}[h]
\begin{tabular}{ll}
\bf 1a. & \tt{\itshape Shanivar neeras hai} from \\ 
& \tt that perspective. \\
\end{tabular}
\end{table}
\vspace*{-\baselineskip} \par

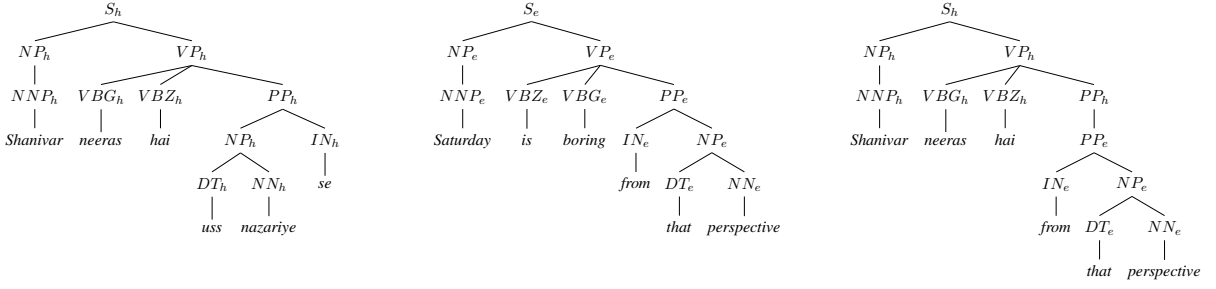
\begin{figure*}
% \begin{adjustbox}{width=\textwidth}
\begin{tikzpicture}[scale=0.55]
\Tree [.$S_h$ [.$NP_h$ [.$NNP_h$ \textit{Shanivar} ]] [.$VP_h$ [.$VBG_h$ \textit{neeras} ] [.$VBZ_h$ \textit{hai} ] [.$PP_h$ [.$NP_h$ [.$DT_h$ \textit{uss} ] [.$NN_h$ \textit{nazariye} ]] [.$IN_h$ \textit{se} ]]]]
\begin{scope}[xshift=10cm]
\Tree [.$S_e$ [.$NP_e$ [.$NNP_e$ \textit{Saturday} ]] [.$VP_e$ [.$VBZ_e$ \textit{is} ] [.$VBG_e$ \textit{boring} ] [.$PP_e$ [.$IN_e$ \textit{from} ] [.$NP_e$ [.$DT_e$ \textit{that} ] [.$NN_e$ \textit{perspective} ]]]]]
\end{scope}
\begin{scope}[xshift=20cm]
\Tree [.$S_h$ [.$NP_h$ [.$NNP_h$ \textit{Shanivar} ]] [.$VP_h$ [.$VBG_h$ \textit{neeras} ] [.$VBZ_h$ \textit{hai} ] [.$PP_h$ [.$PP_e$ [.$IN_e$ \textit{from} ] [.$NP_e$ [.$DT_e$ \textit{that} ] [.$NN_e$ \textit{perspective} ]]]]]]
\end{scope}

\end{tikzpicture}
\caption{Parse-trees of (a) sentences [1H] and (b) [1E], and (c) of [1a] according to the ML model. Note that here {\tt {\itshape neeras}} has been derived from {\em VBG}, though it is actually {\em JJ} for reasons explained in Sec.~\ref{sec:impl}.}
\label{fig:ex1}
% \end{adjustbox}
\end{figure*}

Figure~\ref{fig:ex1} shows the parse trees for the sentences {\bf 1H}, {\bf 1E} and {\bf 1a}. In this case, categorical congruence between the trees is evident from the category names (we follow the Penn Treebank\footnote{http://www.cis.upenn.edu/~treebank/} convention for naming the constituents, subscripted by {\em e}, {\em h} or {\em x} depending on whether they are Hi, En or undetermined/both respectively). 

%However it is not necessarily the case.

% For example,
% \begin{table}[h]
% \begin{tabular}{ll}
% \bf [1b] & \tt Shanivar neeras hai from \\ 
% & \tt that point of view \\
% \end{tabular}
% \end{table}
% \noindent cannot be translated into Hindi exactly, because the English phrase `point of view' has no Hindi equivalent. 

\subsection{The Matrix-Language Theory}
According to the ML theory $G_X$ is the union of two intermediate grammars, $G_{12}$ and $G_{21}$, which generate sentences with $L_1$ and $L_2$ as the {\em matrix language} respectively, such that: 

%\begin{center} $G_X=G_{12} \cup G_{21}$ \end{center}
%where

\begin{center} $G_{12} = \langle V_1\cup V_2, \Sigma_1\cup\Sigma_2, R_1\cup R_2\cup R_{12}, S_1\rangle$ \end{center}
\begin{center} $G_{21} = \langle V_1\cup V_2, \Sigma_1\cup\Sigma_2, R_1\cup R_2\cup R_{21}, S_2\rangle$ \end{center}

\noindent where

\begin{center} $R_{12} = \{v\rightarrow f_{12}(v)|v\in V_1 - \{S_1\} - D_{12}\}$ \end{center}
\begin{center} $R_{21} = \{v\rightarrow f_{21}(v)|v\in V_2 - \{S_2\} - D_{21}\}$ \end{center}

\noindent The extra production rules in $R_{12}$ (and $R_{21}$) essentially allow any category of $L_1$ to be switched by a congruent category of $L_2$. Thus, in example {\bf 1a} (Fig.~\ref{fig:ex1}c), the En constituent {\tt from that perspective} is {\em embedded} in the Hi matrix sentence by allowing the production: $PP_h \rightarrow PP_e$. 
Trivially, $S_1$ and $S_2$ cannot be replaced by each other since that would alter the matrix language of the sentence itself. 

The ML theory issues broad guidelines as to other categories that may not be independently embedded, which are collectively represented as $D_{12}$ and $D_{21}$. One important directive is that categories corresponding to functional units or closed-class items (prepositions, auxilliary verbs, etc.) cannot be replaced by their equivalents in the embedded language. Also note that $R_{12}$ (and $R_{21}$) allows switch from a category in $L_1 (L_2)$ to $L_2 (L_1)$ but not vice versa. This implies that once a category is switched, say the $PP_h \rightarrow PP_e$ in {\bf 1a}, no further switches to the matrix language are allowed within the subtree rooted at $PP_e$. Hence, sentences like {\tt {\itshape Shanivar neeras hai} from that {\itshape nazariya}} are not accepted by the ML model.

\subsection{The Equivalence-Constraint Theory}
The EC theory makes some assumptions that the previous one does not. Namely, it assumes\par
(1) \textit{Lexical congruence}, i.e., in a pair of equivalent sentences $l_1 \in L_1$ and $l_2 \in L_2$, for every lexical unit $w$ in $l_1$, there is an equivalent $u = g(w)$ in $l_2$. \par
(2) \textit{Grammatical congruence} between $G_1$ and $G_2$. For every rule $c_1 \rightarrow v_1 v_2 ... v_n$ in $R_1$, there is exactly one rule $c_2 \rightarrow u_1 u_2 ... u_n$ in $R_2$, such that $c_2=f_{12}(c_1)$ and every $v_i$ has some equivalent $u_j=f_{12}(v_i)$ or $g(v_i)$. This is also a bijection $h:R_1 \rightarrow R_2$.\par

\begin{figure*}
\begin{center}
\begin{tikzpicture}[scale=0.55]

\Tree [.$S_x$ [.$NP_x$ [.$NNP_x$ \textit{Shanivar} ]] [.$VP_x$ [.$VBG_x$ \textit{neeras} ] [.$VBZ_x$ \textit{hai} ] [.$PP_x$ [.$IN_x$ \textit{from} ] [.$NP_x$ [.$DT_x$ \textit{that} ] [.$NN_x$ \textit{perspective} ]] ]]]

\begin{scope}[xshift=12cm]
\Tree [.$S_x$ [.$NP_x$ [.$NNP_h$ \textit{Shanivar} ]] [.$VP_x$ [.$VBG_h$ \textit{neeras} ] [.$VBZ_h$ \textit{hai} ] [.$PP_e$ [.$IN_e$ \textit{from} ] [.$NP_e$ [.$DT_e$ \textit{that} ] [.$NN_e$ \textit{perspective} ]] ]]]
\end{scope}

\end{tikzpicture}
\end{center}
\caption{Parse-trees of sentence [1a] after (a) production stage and (b) verification stage of EC model}
\label{fig:1ec}
\end{figure*}
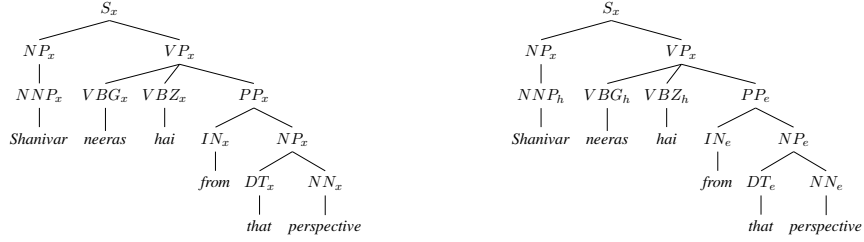

On the basis of these assumptions, the EC theory describes $L_X$ by defining one set of rules to produce a code-switched sentence, and another set to validate the generated sentence. A code-switched sentence $l_X$ is constructed from $l_1$ and $l_2$ following the rules:
(1) For every word $w$ in $l_1$, $l_X$ has either $w$ or $g(w)$ but not both.
(2) A monolingual fragment can occur in $l_X$ only if it also occurs in either $l_1$ or $l_2$.
(3) `Once the production of $l_X$ enters one constituent, it will exhaust all the lexical slots in that constituent or its equivalent constituent in the other language before entering into a higher level consituent or a sister constituent.' ~\cite{sankoff1998}. Essentially, the constituent structure of $l_1$ and $l_2$ will be maintained in $l_X$.\par

Algorithm 4 in supplementary material describes the production process. The reader can verify that {\bf 1a} can be generated from {\bf 1H} and {\bf 1E}. 

 The production process also ensures that the parse-tree $p_X$ of $l_X$ sentence has the same constituent structure as the parse-trees $p_1$ and $p_2$ of $l_1$ and $l_2$ respectively. In fact, $p_X$ must be the product of a grammar $\langle V_X, \Sigma_{1} \cup \Sigma_{2}, R_X, S_X \rangle$, which is categorically and grammatically congruent to $G_1$ and $G_2$. In other words, there are mappings $f_{X}: V_{X} \rightarrow V_1$ and $h_{X}: R_{X} \rightarrow R_1$, similar to the congruence mappings defined earlier, and likewise for $G_2$. Fig.~\ref{fig:1ec}(a) shows the parse-tree for {\bf 1a}.

During the verification stage, each node of $P_X$ is assigned either to $V_1$ or $V_2$, or alternately it remains in $V_X$. This happens in two steps: \par
(1) While conducting a post-order traversal of the parse-tree, if all the children of a node are in either $V_{1} \cup \Sigma_{1}$ or $V_{2} \cup \Sigma_{2}$, the node is assigned to $V_{1}$ or $V_{2}$ respectively. Else, it remains in $V_{X}$. (See algorithm 5 in supplementary material.) Fig.~\ref{fig:1ec}(b) illustrates the parse-tree for {\bf 1a}  after this step.\par
(2) Next, every node is labelled according to the rule applied at its parent node, and its position among its siblings. If the node does not occupy the position dictated by $R_1$, it is assigned to $V_2$, and if it does not occupy the position dictated by $R_2$, it is assigned to $V_1$. The parse-tree of {\bf 1a} is not modified further by this step. If any node in the tree is assigned to both $V_1$ and $V_2$, the sentence is marked invalid and discarded. See algorithm 6 in supplementary material. Otherwise, the assignments made allow the identification of code-switch points in the parse-tree - any location where a node in $V_{1}\cup\Sigma_{1}$ is adjacent to one in $V_{2}\cup\Sigma_{2}$. The only code-switch junction in {\bf 1a} occurs between $VBZ_h$ and $PP_e$.\par 

Finally, the Equivalence Constraint is applied (Algorithm 7 in the supplementary material), which ensures that a code-switch made at one point in a sentence does not necessitate another code-switch at a later point. Let $c_1 \rightarrow v_1 v_2 ... v_n \in R_1$ and $d_1 \rightarrow u_1 u_2 ... u_n \in R_2$ be grammatically congruent. Let, in the generation of $l_X$, \par
\begin{center}$c_X \rightarrow v_1...v_i u_{i+1}...v_n$\end{center}
occur, with a code-switch at the $v_i-u_{i+1}$ junction. This code-switch point satisfies the EC if categorical congruence $h$ maps each category in $v_1...v_i$ to some category in $u_1...u_i$. If every code-switch point in $l_X$ satisfies the constraint, $l_X$ is an acceptable CS sentence according to the EC model.

It is interesting to note that the EC model can also be stated as a context-free grammar $G_X$.

\subsection{Comparing the Models}
\begin{comment}
\begin{figure*}
% \begin{adjustbox}{width=\textwidth}
\begin{tikzpicture}[scale=0.55]
\Tree [.S [.NP [.NNP \textit{Shanivar} ]] [.VP [.VBG \textit{neeras} ] [.VBZ \textit{hai} ] [.PP [.NP [.DT \textit{that} ] [.NN \textit{perspective} ]] [.IN \textit{se} ] ]]]
\begin{scope}[xshift=9cm]
\Tree [.S [.NP [.NNP \textit{Shanivar} ]] [.VP [.VBG \textit{neeras} ] [.VBZ \textit{hai} ] [.PP [.NP [.DT \textit{that} ] [.NN \textit{perspective} ]] [.IN \textit{se} ] ]]]
\end{scope}
\begin{scope}[xshift=18cm]
\Tree[.S [.NP \textit{Main} ] [.VP [.V \textit{kaam karta hoon} ] [.PP [.IN \textit{in} ] [.NP [.DT \textit{the} ] [.ADJP \textit{ghar ke neeche wala} ] [.NN \textit{dafter} ]]]]]
\end{scope}
\end{tikzpicture}
\caption{Parse-trees of sentences (a) [1b] and (b) [1c], and (c) [2a]}
% \end{adjustbox}
\end{figure*}
\end{comment}

Which of these models explains the phenomenon of CS better? To answer this question, we would need to verify whether all the CS sentences generated by a model are acceptable, i.e., whether the model is {\em sound}, and whether all acceptable CS sentences are generated by the model, i.e., whether the model is {\em complete}. This is practically impossible to do as (a) the set of possible CS sentences for even a pair of equivalent sentences $l_1$ and $l_2$ is very large, and (b) acceptability is a relative notion that would require a large scale user study. 

We therefore propose the notions of theoretical and empirical equivalence between the two models. For a given pair of $l_1$ and $l_2$, let $L^{ML}_X(l_1,l_2)$ and $L^{EC}_X(l_1,l_2)$ be the sets of CS sentences accepted/generated by the ML and EC models respectively. 
The two models are said to be {\bf theoretically equivalent} if $L^{ML}_X(l_1,l_2) = L^{EC}_X(l_1,l_2)$ for every pair $l_1, l_2$. We say EC {\em subsumes} ML if and only if $L^{ML}_X(l_1,l_2) \subset L^{EC}_X(l_1,l_2)$, and vice versa.
The models are {\bf empirically equivalent}, for a given set of $l_1, l_2$ pairs, if the number of sentences judged acceptable in $L^{ML}_X(l_1,l_2) - L^{EC}_X(l_1,l_2)$ by bilingual speakers is same as the number of acceptable sentences in $L^{EC}_X(l_1,l_2) - L^{ML}_X(l_1,l_2)$. Here `$-$' represents set difference.

It is evident that $L^{EC}_X(l_1,l_2) \cap L^{ML}_X(l_1,l_2)$ is non-null for most $l_1,l_2$ (e.g., {\bf 1a} is accepted by both the models). However, the models are neither theoretically equivalent, nor does one of them subsume the other. Consider examples {\bf 1b} and {\bf 1c}.\\

\vspace*{-\baselineskip}
\begin{table}[H]
\begin{tabular}{ll}
\bf 1b. & \tt {\itshape Shanivar neeras hai} that \\ 
& \tt perspective {\itshape se}\\
\end{tabular}
\end{table}
\vspace*{-\baselineskip} \par
\vspace*{-0.5\baselineskip}
\begin{table}[h]
\begin{tabular}{ll}
\bf 1c. & \tt {\itshape Shanivar neeras hai} that \\
& \tt {\itshape nazariye se}
\end{tabular}
\end{table}
\vspace*{-\baselineskip} \par 

It is easy to see that {\bf 1b} is accepted by the ML model, but {\bf 1c} is not because in the ML model, a functional category, in this case the determiner {\em that}, cannot be switched in isolation. On the other hand, the EC model accepts {\bf 1c} but rejects {\bf 1b}. This is because the noun-phrase \textit{that perspective} is identified as a category of En. However, as it does not occupy the second position among its sibling nodes, as required by the En rule ($PP_{e} \rightarrow IN_{e}$ $NP_{e}$). So, it is also identified as a category of Hi. Due to this clash, the EC model rejects the sentence.

In order to empirically compare the two models, we will have to implement the models and generate sentences to be judged by speakers. The next two sections describe the implementation and experiments.

%% file: 3.implementation.tex
\section{Implementation of the Models}
\label{sec:impl}
Ideally, one should be able to take a sentence in $L_1$, automatically translate it to $L_2$ using a machine translation system, automatically align the two sentences at word-level and use parsers of $L_1$ and $L_2$ to parse the sentences, after which both ML and EC models can be run on the aligned parse-trees. 

However, our initial attempt at this approach failed drastically because both the models require very accurate and literally translated pairs as inputs, which the current machine translation systems are not able to produce\footnote{We experimented with Google and Bing MT Systems for En-Hi, which are the best available translators for these languages. However, even for simple and short sentences, the system translations did not serve the purpose.}. Further, we experimented with the Berkeley aligner\footnote{https://code.google.com/archive/p/berkeleyaligner/} ~\cite{berkeley-parser-1,berkeley-parser-2}, however, the models were very sensitive to even small alignments errors (accurate alignments were also necessary for correct projection of the En parse-trees on the Hi side, as there are no high accuracy parsers for Hi). Due to paucity of space we do not report these experiments. Instead, here we shall assume that the input to the systems is a pair of accurately translated sentences $l_1$ and $l_2$ along with correct word-level alignments. 

A second set of challenges arises due to the underspecification of the original models. As we shall describe in Secs~\ref{sec:const} and \ref{sec:modify}, we make necessary assumptions and systematically modify them in a manner that enables the models to achieve their best possible performance on real data.

We describe our implementation of the models for En-Hi, though the implementation is language-independent except for the  language-specific parser.

\subsection{Parsing}
As we have seen in Sec. 2, both ML and EC models require the parse-tree of the pair of input sentences. We use the Stanford Parser\footnote{http://nlp.stanford.edu:8080/parser/} ~\cite{klein-stanford-parser} to parse En sentences. Since there is no equivalent parser for Hindi, we project the Hindi parse-tree from the English parse-tree using the word-level alignments.
The projection works bottom up as follows: Let according to the parse tree of $l_1$ (here in En), $v \in V_1$ produce the words $w_i w_{i+1}...w_{i+k}$. If the words $g(w_i), g(w_{i+1}), ..., g(w_{i+k})$ occur as a contiguous fragment in $l_2$ (here in Hi), then we introduce a node $f_{12}(v)$ in the parse tree of $l_2$ and make  $g(w_i), g(w_{i+1}), ..., g(w_{i+k})$ its children. This process is followed recursively, and stops at $S_2$. 

\begin{figure*}
% \begin{adjustbox}{width=\textwidth}

\begin{tikzpicture}[scale=0.55]
\Tree[.S [.NP [.NP [.PRP \textit{Our} ] [.NN \textit{chance} ]] [.PP [.IN \textit{of} ] [.NP [.NP [.NN \textit{survival} ]] [.PP [.IN \textit{in} ] [.NP [.DT \textit{this} ] [.NN \textit{war} ] ]]]]] [.VP [.VBZ \textit{is} ] [.ADJP [.JJ \textit{low} ]]]]
\begin{comment}
\begin{scope}[xshift=11cm]
\Tree[.S [.NP [.$PP_1$  [.$NP_2$ [.PP  [.NP [.DT \textit{Iss} ] [.NN \textit{jung} ]] [.IN \textit{mein} ]] [.NP [.NN \textit{bachne} ]] ] [.$IN_1$ \textit{ki} ]] [.$NP_1$ [.PRP \textit{hamare} ] [.$NN_1$ \textit{sambhavna} ]]] [.VP [.ADJP [.JJ \textit{kam} ]] [.VBZ \textit{hai} ]]]
\end{scope}
\end{comment}
\begin{scope}[xshift=10cm]
\Tree[.S [.NP [.PRP \textit{Our} ] [.NN \textit{chance} ] [.IN \textit{of} ] [.NP [.NN \textit{survival} ]] [.PP [.IN \textit{in} ] [.NP [.DT \textit{this} ] [.NN \textit{war} ] ] ]    ] [.VP [.VBZ \textit{is} ] [.ADJP [.JJ \textit{low} ]] ]]
\end{scope}
\begin{scope}[xshift=20cm,yshift=-2cm]
\Tree[.S [.NP [.PP  [.NP [.DT \textit{Iss} ] [.NN \textit{jung} ] ] [.IN \textit{mein} ]] [.PRP \textit{hamare} ] [.NP [.NN \textit{bachne} ]] [.IN \textit{ki} ] [.NN \textit{sambhavna} ]] [.VP [.ADJP [.JJ \textit{kam} ]] [.VBZ \textit{hai} ]]]
\end{scope}
\end{tikzpicture}
\caption{(a) Parse-tree of sentence {\bf 2E}, (b) modified parse-tree of {\bf 2E} and (c) projected and modified parse-tree of {\bf 2H}.}
% \end{adjustbox}Parse-trees of sentences 
\label{fig:projection}
\end{figure*}
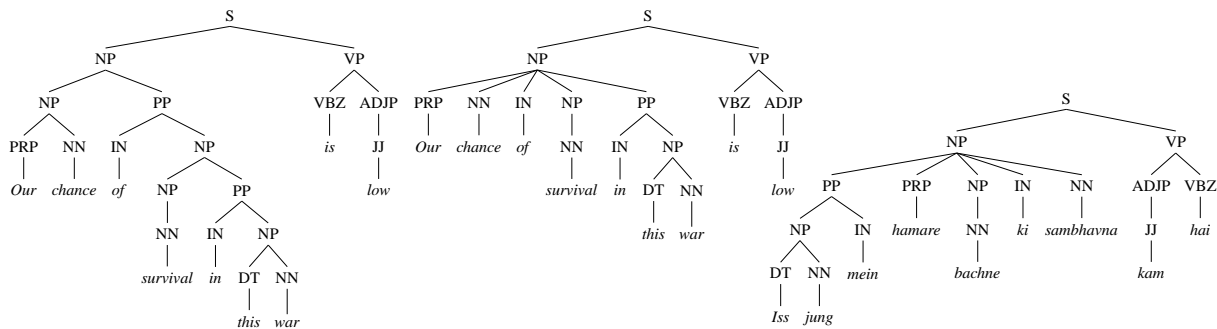

In cases where the words are non-contiguous in $l_2$, no node is created corresponding to $v$; creation of a node is deferred till a node which is an ancestor of $v$ is found, for which the above condition is met. This is illustrated in Fig.~\ref{fig:projection} for {\bf 2E} and {\bf 2H}.

\vspace*{-\baselineskip}
\begin{table}[H]
\begin{tabular}{lccccc}
\bf 2H. & \tt\itshape Iss & \tt\itshape jung & \tt\itshape mein & \tt\itshape hamare& \tt\itshape bachne \\ 
& This & war & in & our & survival\\ 
\end{tabular}
\end{table}
\vspace*{-\baselineskip} \par
\vspace*{-\baselineskip}
\begin{table}[H]
\begin{tabular}{p{4mm}cccc}
\vspace*{-\baselineskip}
  & \tt\itshape ki &  \tt\itshape sambhavna &\tt\itshape kam &\tt\itshape hai.  \\
  & of & chance & low & is \\
\end{tabular}
\end{table}
\vspace*{-\baselineskip} \par

\vspace*{-\baselineskip}
\begin{table}[H]
\begin{tabular}{ll}
\bf 2E. & \tt Our chance of survival in this\\ 
& \tt  war is low.\\
\end{tabular}
\end{table}
\vspace*{-\baselineskip} \par 

The English parse-tree is also modified alongside so that the constituent structures of the two trees are symmetric. Finally, grammar rules of both languages are inferred from the parse-trees. Since the Hi tree is a projection of the English tree, categorical and grammatical congruence assumptions are upheld. However, the alteration of the parse-trees results in rules and phrase-structures that do not map exactly to the those of the natural language grammars.

\subsection{Modelling Constraints}
\label{sec:const}
Once the parse trees are generated, the models described in Sec 2 (and also as algorithms 1, 4, 5, 6, and 7 in the supplemenary material) can be readily applied to generate the set of code-switched sentences for the models. 

The one major challenge in using the ML model is that its constraints are highly underspecified. It does not exhaustively list or describe the categories of the matrix language that cannot be replaced by their counterparts in the embedded language. While one constraint explicitly states that in the main verb phrase of a sentence, any auxilliary/helping verb and verbs in tense may not be swapped, another disallows the replacement of 'closed class items', which we interpret as forbiddance of the swapping of any purely functional category. Currently, we do not implement any constraints that address very specific circumstances, such as those regarding complimentizers.\par

The model explains that functional categories (such as a pronoun or preposition) can be code-switched as a part of another category (say, a noun phrase or prepositional phrase), but not in isolation. However, it does not discuss the status of a category that is comprised only of categories that may not be switched independently (for example an $NP$ that derives a pronoun and adposition).  We disallow the substitution of such categories. 

% \begin{center}
% \begin{figure}
% \begin{tikzpicture}[scale=0.55]
% \Tree[.S [.NP [.PRP \textit{I} ]] [.VP [.VBD \textit{waved} ] [.PP [.TO \textit{to} ] [.NP [.PRP \textit{them} ]]]]]
% \end{tikzpicture}
% \end{figure}
% \end{center}

The EC model is relatively more straightforward to implement, since it does not make any distinction between different categories or grammatical rules.

\subsection{Modification of Constraints on the Models}
\label{sec:modify}

Real Hi-En CS data from social media shows that both the models are highly constrained and do not allow some commonly seen CS patterns. We remove such constraints that are forbidding these observed patterns, without modifying the essential core of the models. For ease of reference, let us denote the systems that faithfully model the EC and ML theories as described in ~\cite{sankoff1998} and \cite{Joshi85} as $EC_0$ and $ML_0$ respectively. 

\subsubsection{Lexical Substitution and Well-Formed Fragments}

The EC model, unlike the ML model, does not account for lexical substitution (those are essentially explained away as borrowing, which is debatable), and hence does not generate a large number of valid sentences. For instance, {\bf 2a} is not accepted by the EC model, because \textit{sambhavna} is a Hi word which does not occupy the position in the parse-tree that the Hi grammar requires of it.

\vspace*{-\baselineskip}
\begin{table}[H]
\begin{tabular}{ll}
\bf 2a. & \tt Our {\itshape sambhavna} of survival\\ 
& \tt in this war  is low.\\
\bf 2b. & \tt{\itshape Iss jung mein hamare bachne} \\ 
& \tt {\itshape ki} chance low {\itshape hai.}\\
\end{tabular}
\end{table}
\vspace*{-\baselineskip}

We define a new model $EC_1$ which allows lexical substitution of all nouns, adjectives and other such content-only-lexemes. The sentences involving lexical substitution that are produced at the generation stage of the EC model are not discarded at the verification stage. Note that sentences such as {\bf 2b} are not even produced at the generation stage of the model, since the model disallows ill-formed monolingual fragments (the sequence \textit{chance low} does not occur in the English sentence). 

\subsubsection{Nested Switching and Ill-Formed Fragments}

Recall that the ML model does not allow switching back to the matrix language, once a category is switched to the embedded language. We relax this condition in model $ML_1$. Note that this is a significant departure from the original ML model. $ML_1$ generates an exponential number of sentences in terms of the number of switchable categories in the parse tree. Such a model is also expected to generate many sentences, where monolingual fragments are ill-formed. Therefore, we further introduce $ML_2$, which is obtained by eliminating all the sentences with ill-formed monolingual segments from $ML_1$.

%% file: 4.emperical_analysis.tex
\section{Empirical Analysis of Models}
The number of CS sentences generated by the models for a pair of input sentences varied from around ten to a few thousands. Therefore, an empirical study with even a few sentences would be quite challenging. Hence, for this work, we settle for a small scale empirical study, which we deem more as a pilot.

\subsection{Basic Comparison}
We collected 25 English sentences and 25 Hindi sentences from the Internet (including BBC News, Dainik Bhaskar News\footnote{http://www.bhaskar.com/}, a culinary blog\footnote{http://nishamadhulika.com/en/}, Twitter\footnote{https://twitter.com/?lang=en}), with an average of 11.8 words per sentence. This set was chosen to contain a mix of commonly used sentence structures, tense and voice. The En (Hi) sentences were translated to Hi (En) by a linguist fluent in both the languages,  maintaining syntactic and lexical congruence as much as possible. Then the words were aligned by the same expert, and then given as input to the EC and ML modules after parsing and projection.

\begin{table*}
\begin{center}
\begin{tabular}{|c|c|c|c|c|c|c|c|c|c|}
\hline \bf Sent- & \bf Length & \bf Cate-& \bf Depth & \bf BF  & \bf $EC_0$ & \bf $EC_1$ & \bf $ML_0$ & \bf $ML_1$ & \bf $ML_2$ \\
\bf ence & & \bf gories &  & & & \bf & \bf $(\cap EC_{1})$ & \bf $(\cap EC_{1})$ & \bf $(\cap EC_{1})$ \\ \hline
1 & 6 & 11 & 4 & 3 & 22 & 22 & 8(6) & 32(22) & 28(22) \\
2 & 9 & 16 & 4 & 5 & 14 & 44 & 34(18) & 256(44) & 132(44)  \\
3 & 5 & 10 & 3 & 3 & 16 & 28 & 16(2) & 32(28) & 28(28) \\

4 & 9 & 18 & 5 & 3 & 48 & 56 & 16(8) & 64(16) & 40(16)\\
5 & 12 & 25 & 6 & 3 & 96 & 112 & 28(20) & 256(112) & 208(112)\\
\hline
\end{tabular}
\end{center}
\caption{ Experimental Results. Columns from left to right: Sentence index, number of words, number of categories after parse-tree modification, depth of parse-tree, maximum branching factor in tree, number of sentences generated by $EC_0$, $EC_1$, $ML_0$, $ML_1$ and $ML_2$ systems. In columns with results of ML systems, value in brackets indicate number of sentences in common with $EC_1$.} 
\end{table*}

 Table 1 shows the numbers of CS sentences generated by the various systems for sentence pairs {\bf 1H-1E}, {\bf 2H-2E} and the following sentences (For paucity of space, we show only the En versions. See supplementary material for the corresponding Hi versions).

\begin{comment}

\begin{figure*}[h]
\begin{center}
\begin{tikzpicture}[scale=0.55]
\begin{scope}
\Tree[.S [.ADVP [.RB \textit{Now} ]] [.VP [.VP [.VB \textit{pour} ] [.NP [.NN \textit{oil} ]] [.PP [.IN \textit{in} ] [.NP [.DT \textit{the} ] [.NN \textit{pan} ]]]] [.CC \textit{and} ] [.VP [.VB \textit{heat} ] [.NP [.PRP \textit{it} ]]]]]
\end{scope}
\begin{scope}[xshift=13cm]
\Tree[.S [.SBAR [.IN \textit{upon} ] [.VP [.VBG \textit{knowing} ] [.NP [.DT \textit{this} ]]]] [.S [.NP [.NP [.NNS \textit{people} ]] [.PP [.RB \textit{instead} ] [.IN \textit{of} ] [.VP [.VBG \textit{helping} ] [.NP [.DT \textit{the} ] [.NN \textit{driver} ]]]]] [.VP [.VBD \textit{started} ] [.VP [.VBG \textit{stealing} ] [.NP [.NNS \textit{bananas} ]]]]]]
\end{scope}
\end{tikzpicture}
\end{center}
\caption{Parse trees of sentences (a) {\bf 5E} and {\bf 6E}}
\end{figure*}
\end{comment}

\begin{comment}
\vspace*{-0.5\baselineskip}
\begin{table}[H]
\begin{tabular}{lcccc}
\bf 3H. & \tt\itshape Kuch & \tt\itshape log & \tt\itshape sirf \\ 
& Some & people & only \\ 
  & \tt\itshape samay & \tt\itshape barbaad & \tt\itshape karte hain \\
 & time & waste & (aux. verb)
\end{tabular}
\end{table}
\vspace*{-\baselineskip} \par 
\end{comment}
\vspace*{-0.5\baselineskip}
\begin{table}[H]
\begin{tabular}{ll}
\bf 3E. & \tt Some people only waste time \\
\end{tabular}
\end{table}
\vspace*{-\baselineskip} \par 
\begin{comment}
\vspace*{-0.5\baselineskip}
\begin{table}[H]
\begin{tabular}{lllllll}
\bf 4H. & \tt {\itshape Ab} & \tt {\itshape kadhai} & \tt {\itshape mein} & \tt {\itshape tel} \\
& Now & pan & in  & oil\\
 & \tt {\itshape dalo} & \tt {\itshape aur} & \tt {\itshape usko} \\ 
& pour & and & it \\
& \tt {\itshape garam karo}\\
& heat
\end{tabular}
\end{table}
\vspace*{-\baselineskip}
\end{comment}
\vspace*{-0.5\baselineskip}
\begin{table}[H]
\begin{tabular}{lllllll}
\bf 4E. & \tt {Now pour oil in the pan and}\\ 
& \tt{heat it}
\end{tabular}
\end{table}
\vspace*{-\baselineskip}
\begin{comment}
\vspace*{-0.5\baselineskip}
\begin{table}[H]
\begin{tabular}{lllllll}
\bf 5H. & \tt {\itshape Iska} & \tt {\itshape pata} & \tt {\itshape chalte hi} \\
& This & knowing & upon\\
 & \tt {\itshape logon ne} & \tt {\itshape bajay} & \tt {null} \\ 
& people & instead & the \\
 & \tt {\itshape driver ki} & \tt {\itshape madad karne} & \tt {\itshape ke} \\ 
& driver & helping & of \\
 & \tt {\itshape kele} & \tt {\itshape lootna} & \tt {\itshape shuru kiya} \\ 
& bananas & stealing & started \\
\end{tabular}
\end{table}
\vspace*{-\baselineskip}
\end{comment}
\vspace*{-0.5\baselineskip}
\begin{table}[H]
\begin{tabular}{lllllll}
\bf 5E. & \tt {Upon knowing this people}\\ 
& \tt{instead of helping the}\\
& \tt{driver started stealing}\\
& \tt{bananas}
\end{tabular}
\end{table}
\vspace*{-\baselineskip}
We note that as expected, the relaxation and imposition of constraints do shrink and expand the sets of sentences generated. The EC systems generate more sentences for sentence-pair 4 then for sentence-pair 2. This is possibly because code-switching is extremely constrained at the node with branching factor 5 in sentence-pair 2, with many arrangements of words being discarded at the verification phase of the EC system. We note that a significant number of sentences generated by $ML_1$ are ill-formed, and hence duly discarded in $ML_2$. 

Interestingly, the set of sentences generated by $ML_2$ subsumes that of $EC_1$ for all but sentence pair 4. This is because of elements such as the conjunction \texttt{and}, which the EC model can code-switch, but the ML model cannot.

\subsection{Human Evaluation}

\begin{table*}
\footnotesize
\begin{center}
\begin{tabular}{|l|c|c|c|c|}
\hline \bf Sentence & \bf Min & \bf Max & \bf Average & \bf $\sigma$ \\ \hline
\tt{Saturday is boring \textit{uss nazariye se}} & 1(3) & 5(5) & 3.6(4) & 1.67(1) \\
\tt{\textit{Shanivar neeras hai} from that point of view}	& 1(1) & 4(4) & 2.4(3) &	1.52(1.22) \\
\tt{Curfew has been imposed \textit{saat se zyada din se}} & 3(3) & 5(5) & 3.8(4) & 0.84(0.71) \\
\tt{Any \textit{chattra isse samaj sakta hai}} & 3(2) & 4(4) & 3.2(2.8) & 0.45(1.09) \\
\tt{Will \textit{tum} come tomorrow?}	& 1(1) & 3(4) & 1.8(2.2) & 0.84(1.09) \\
 \hline
\end{tabular}
\end{center}
\caption{Human Judgement of Usability and Fluency (in parenthesis)}
\end{table*}

% \begin{table*}
% \footnotesize
% \begin{center}
% \begin{tabular}{|l|c|c|c|c|}
% \hline \bf Sentence & \bf Min & \bf Max & \bf Average & \bf $\sigma$ \\ \hline
% \tt{Saturday is boring \textit{uss nazariye se}} & 3 &	5 &	4 & 1 \\
% \tt{\textit{Shanivar neeras hai} from that point of view}	& 1	&4	&3	&1.22 \\
% \tt{Curfew has been imposed \textit{saat se zyada din se}} & 3 & 5 & 4 & 0.71 \\
% \tt{Any \textit{chattra isse samaj sakta hai}} & 2 & 4 & 2.8 & 1.09 \\
% \tt{Will \textit{tum} come tomorrow?}	& 1 & 4 & 2.2 & 1.09 \\
%  \hline
% \end{tabular}
% \end{center}
% \caption{Human Judgement of Fluency}
% \end{table*}

In order to judge the acceptability of the generated sentences, we asked five fluent En-Hi bilinguals to judge a set of CS sentences generated by the models on two criteria: likelihood of usage (0 - ``do not expect to hear or use ever'' to 5 - ``extremely likely to be heard/used'') and fluency (0 - ``Does not make sense or sound right" to 5 - ``absolutely clear and fluent"). The judges were all native Hi speakers who has acquired En either along with Hi or a few years later, but were certainly fluent bilinguals by the age of 8. Their ages ranged from 22 to 24 years, they all had an undergraduate degree and grew up in different Indian cities. 

The sentences were chosen based on some interesting phenomena (e.g., switching of a pronoun or functional category, or nested switching) that we were interested in investigating. Table 2 reports aggregate statistics on the judgment scores for 5 sentences. Except for the third sentence, there is a large variation in the scores of the 5 judges, which is evident from the min-max values and the standard deviation. This implies that acceptability thresholds for CS sentences vary widely in the bilingual population and might be influenced by various socio-cultural and region specific factors. Nevertheless, we do observe trends, for instance, the switching of a pronoun (in the fifth example, {\tt tum} = you) seems generally unacceptable as is suggested by the ML model.

It is difficult to comment on the relative performance of the two models, but even $EC_1$ seems to be more constrained than necessary for Hi-En CS.

%% file: 5._discussion_litsurvey.tex
\section{Discussion and Conclusion}
In this paper, we implemented two popular linguistic theories of intra-sentential code-switching. While these theories have existed for more than three decades and have been discussed, debated and modified actively, we do not know of any earlier attempts to design computational systems based on these models that could accept and/or generate CS sentences. 

There are several important insights that we gained from this study. First, the theories are underspecified and one has to make several non-trivial assumptions while building the models, which determine the performance. Second, neither of the models is sound or complete, nor does one subsume the other. Third, acceptability of CS patterns is influenced as much if not more by socio-linguistic factors as by cognitive factors. This is evident not only from the large variance observed in the human evaluation, but also from the fact that the $ML_0$ model proposed by Joshi in 1985 for En and Marathi seems to be much more constrained than what current Hi-En CS patterns indicate. It is not unreasonable to assume that the constraints such as 'functional categories cannot be switched in isolation' were deduced based on the Marathi-En CS patterns in 1985. Since current Hi-En bilinguals seem to accept switching of several (but not all) of the functional categories, we wonder if the constraints on CS are progressively getting weaker. 

   An important aspect of acceptabilty, especially fluency and naturalness judgment, is dependent on word collocations and other lexical factors. For instance, some of the human judges pointed out that words such as {\tt neeras} are too formal to be used in a CS sentence. In other words, there seems to be a notion of registers within each language and speakers code-switch only or primarily in informal situations and therefore, use words from the informal or low registers while code-switching. We believe this is an important aspect of CS, which can even make switching obligatory, at least at the lexical level, under certain situations. 
  
  In conclusion, the definition of grammatical models of intra-sentential CS is far from a solved problem, both in theory and in practice. There are several open problems. In particular, it seems that there is a series of constraints, probably a hierarchy or partial-order, on CS and each bilingual community has its own threshold in this hierarchy which defines what is an acceptable CS sentence to a community of speakers. Moreover, the threshold also seems to evolve over time, changing the patterns of CS between two languages. We believe that it is a very interesting research agenda to formulate the set of constraints and their ordering relations.

%% file: glosses.tex
\section{Omitted Sentences}

Some English sentences introduced in Section 4.1 were not accompanied by their Hindi versions. The corresponding Hindi sentences are:

\vspace*{-0.5\baselineskip}
\begin{table}[H]
\begin{tabular}{lcccc}
\bf 3H. & \tt\itshape Kuch & \tt\itshape log & \tt\itshape sirf \\ 
& Some & people & only \\ 
  & \tt\itshape samay & \tt\itshape barbaad & \tt\itshape karte hain \\
 & time & waste & (aux. verb)
\end{tabular}
\end{table}
\vspace*{-\baselineskip} \par 

\vspace*{-0.5\baselineskip}
\begin{table}[H]
\begin{tabular}{lcccccc}
\bf 4H. & \tt {\itshape Ab} & \tt {\itshape kadhai} & \tt {\itshape mein}  \\
& Now & pan & in  \\
 & \tt {\itshape tel} & \tt {\itshape dalo} & \tt {\itshape aur}  \\ 
& oil & pour & and  \\
& \tt {\itshape usko} & \tt {\itshape garam karo}\\
& it & heat
\end{tabular}
\end{table}
\vspace*{-\baselineskip}

\vspace*{-0.5\baselineskip}
\begin{table}[H]
\begin{tabular}{lcccccc}
\bf 5H. & \tt {\itshape Iska} & \tt {\itshape pata}  \\
& This & knowing \\
& \tt {\itshape chalte hi} & \tt {\itshape logon ne}  \\ 
& upon & people  \\
& \tt {\itshape bajay} & \tt {-null-} \\
& instead & the \\
 & \tt {\itshape driver ki} & \tt {\itshape madad karne} \\ 
& driver & helping  \\
& \tt {\itshape ke}  & \tt {\itshape kele} \\ 
& of & bananas \\
& \tt {\itshape lootna} & \tt {\itshape shuru kiya} \\
 & stealing & started \\
\end{tabular}
\end{table}
\vspace*{-\baselineskip}

Here are the monolingual Hindi and English versions of the last three sentences introduced in Table 2 (the first two are code-mixed versions of [1H] and [1E]):

\vspace*{-0.5\baselineskip}
\begin{table}[H]
\begin{tabular}{lcccccc}
\bf 6H. & \tt {\itshape Nishedhagya} & \tt {\itshape lagaya gaya hai}  \\
& Curfew & has been imposed  \\
 & \tt {\itshape saat} & \tt {\itshape se} \\ 
& seventy & than \\
& \tt {\itshape zyada} & \tt {\itshape din}\\
& more & days \\
& \tt {\itshape se} \\
& for
\end{tabular}
\end{table}
\vspace*{-\baselineskip}

\vspace*{-0.5\baselineskip}
\begin{table}[H]
\begin{tabular}{ll}
\bf 6E. & \tt Curfew has been imposed\\
& \tt for more than seventy days \\
\end{tabular}
\end{table}
\vspace*{-\baselineskip}

\vspace*{-0.5\baselineskip}
\begin{table}[H]
\begin{tabular}{lcccccc}
\bf 7H. & \tt {\itshape Koi} & \tt {\itshape chattra} & \tt {\itshape isse}  \\
& Any & student & this  \\
 & \tt {\itshape samaj} & \tt {\itshape sakata hai} &\\ 
& understand & can \\
\end{tabular}
\end{table}
\vspace*{-\baselineskip}

\vspace*{-0.5\baselineskip}
\begin{table}[H]
\begin{tabular}{ll}
\bf 7E. & \tt Any student can understand\\
& \tt this \\
\end{tabular}
\end{table}
\vspace*{-\baselineskip}

\vspace*{-0.5\baselineskip}
\begin{table}[H]
\begin{tabular}{lcccccc}
\bf 8H. & \tt {\itshape Kya} & \tt {\itshape tum} & \tt {\itshape kal}  \\
& -null- & you & tomorrow  \\
 & \tt {\itshape aaoge} & \tt ? \\ 
& will come & ?\\
\end{tabular}
\end{table}
\vspace*{-\baselineskip}

\vspace*{-0.5\baselineskip}
\begin{table}[H]
\begin{tabular}{ll}
\bf 8E. & \tt Will you come tomorrow?\\
\end{tabular}
\end{table}
\vspace*{-\baselineskip}

\vfill

%% file: algorithms.tex
\section{Algorithms}

This supplementary section contains pseudocode for algorithms discussed throughout the paper.

{\bf Algorithm 1} is a recursive method for generating all code-mixed sentences allowed by the Language-Matrix model. The initial arguements passed to the function are (1) the parse-tree of the sentence in the matrix language, (2) the list of categories of the matrix language that may be replaced by their congruent categories in the embedded language and (4) a list containing the root of the parse-tree passed as argument 1. Each node in the parse-tree passed as parameter must have a field pointing to its congruent node in the other parse-tree. All generated code-mixed parse-trees are added to the global set LMSentences.

{\bf Algorithm 2} takes the same arguments as Algorithm 1. It generates sentences according to the LM model, after relaxing the constraint on nested code-switching.

\textbf{Algorithm 3} checks for the well-formedness of monolingual fragments in a code-switched sentence, given the sentence and the monolingual sentences it is a derivative of.

{\bf Algorithm 4} produces sentences according to the generative rules in the EC model, {\bf Algorithms 5} and {\bf 6} assign languages to the categories in the parse-trees of the sentences generated, and {\bf Algorithm 7} checks the parse-trees for the Equivalence Constraint. Arguments to Algorithm 4 are the code-mixed sentence (initially empty), both monolingual sentences and the current position in the parse-tree (initially at the root). The only parameter passed to Algorithm 5 is the root of the code-switched parse-tree, and Algorithms 6 and 7 take this root and the code-mixed sentence as their arguments.

{\bf Algorithm 8} is used to allow lexical substitution in the EC model. It takes the root of the code-mixed sentence and a list of categories that can be lexically substituted as arguments. A tree must be passed to this algorithm only if it fails Algorithm 5.

\begin{algorithm}
\caption{generateLM0($tree, allowList, queue$)}
\begin{algorithmic}
\IF{$queue.empty()$}
	\STATE $LMSentences.add(tree)$
\ELSE
	\STATE $curr=queue.top()$
	\IF{$curr.category$ in $allowList$}
    	\STATE $treeCopy=copy(tree)$
        \STATE $treeCopy.replaceSubtree(curr, curr.counterpart$)
        \STATE $LMSentences.add(treeCopy)$
    \ENDIF
    \FOR{$child$ in $curr.children$}
    	\STATE $queue.pushBack(child)$
    \ENDFOR
    \STATE $generateLM(tree, allowList, queue)$
\ENDIF
\end{algorithmic}
\end{algorithm}

\begin{algorithm}[H]
\caption{generateLM1($tree, allowList, queue$)}
\begin{algorithmic}
\IF{$queue.empty()$}
	\STATE $LMSentences.add(tree)$
\ELSE
	\STATE $curr=queue.top()$
	\IF{$curr.category$ in $allowList$}
    	\STATE $treeCopy=copy(tree)$
        \STATE $queueCopy=copy(queue)$
        \STATE $treeCopy.replaceSubtree(curr, curr.counterpart$)
        \FOR{$child$ in $curr.counterpart.children$}
    		\STATE $queueCopy.pushBack(child)$
    	\ENDFOR
        \STATE $generateLM(treeCopy, allowList, queueCopy)$
    \ENDIF
    \FOR{$child$ in $curr.children$}
    	\STATE $queue.pushBack(child)$
    \ENDFOR
    \STATE $generateLM(tree, allowList, queue)$
\ENDIF
\end{algorithmic}
\end{algorithm}

\begin{algorithm}
\caption{checkFormation($l_{X}, l_{1}, l{2}$)}
\begin{algorithmic}
\FOR {$i$ in $range(0..len(l_{X})-1)$}
	\IF {$l_{X}[i]$ in $l_{1}$ and $l_{X}[i+1]$ in $l_{1}$}
    	\IF{$L_{1}.index(l_{X}[i+1])-L_{1}.index(l_{X}[i]) \neq 1$}
        	\RETURN{False}
        \ENDIF
     \ELSE
     	\STATE ...
     \ENDIF
\ENDFOR
\RETURN{True}
\end{algorithmic}
\end{algorithm}

\begin{algorithm}
\caption{generateECSentences($l_X,l_{1},l_{2},path_{curr}$)}
\begin{algorithmic}
\algsetup{linenosize=\verysmall}
\IF {$l_{X}.empty()$}
	\FOR {$w \in l_1$}
		\STATE $l_{new}=newSentence().insert(w)$
        \STATE $path_{curr} \gets updatePath(path_{curr},w)$
        \STATE $generateECSentences(l_{new},l_{1},l_{2},path_{curr})$
    \ENDFOR
    \FOR {$w \in l_2$}
    	\STATE ...
    \ENDFOR
\ELSIF{$l_{X}.length \neq l_{1}.length$}
	\IF {$l_{X}.lastWord.lang=L_{1}$}
	    \IF {$l_{X}.lastWord \neq l_{1}.lastWord$}
	    	\STATE $index_{curr} \gets l_{1}.index(l_{X}.lastWord)$
	    	\STATE $w=l_{1}[index_{curr}+1]$
	    	\IF {$unused(w)$ and $belongs(path_{curr},w)$}
	    		\STATE $l_{new}=copy(l_{X}).insert(w)$
        		\STATE $path_{curr} \gets updatePath(path_{curr},w)$
        		\STATE $generateECSentences(l_{new},l_{1},l_{2},path_{curr})$
        	\ENDIF
	    \ENDIF
		\FOR {$w \in l_2$}
	  		\IF {$unused(w)$ and $belongs(path_{curr},w)$}
	  			\STATE $l_{new}=newSentence().insert(w)$
       			\STATE $path_{curr} \gets updatePath(path_{curr},w)$
       			\STATE $generateECSentences(l_{new},l_{1},l_{2},path_{curr})$
       		\ENDIF
    	\ENDFOR
    \ELSE
    	\STATE ...
    \ENDIF
\ELSE
   	\STATE $EMSentences.add(l_{X})$
\ENDIF
\RETURN
\end{algorithmic}
\end{algorithm}

\begin{algorithm}
\caption{assignLanguageToCategory($treeNode$)}
\begin{algorithmic}
\IF{$treeNode \in \Sigma_1$ or $treeNode \in \Sigma_2$}
	\RETURN
\ELSE
	\FOR{$c \in treeNode.children$}
		\STATE $assignLanguageToCategory(c)$
	\ENDFOR
	\IF{$c \in V_{1} \cup \Sigma_{1} \forall c \in treeNode.children$}
		\STATE $replace(treeNode,f_{X}(treeNode))$
	\ELSIF{$c \in V_{2} \cup \Sigma_{2} \forall c \in treeNode.children$}
		\STATE $replace(treeNode,f_{12}(f_{X}(treeNode)))$
	\ENDIF
\ENDIF
\RETURN
\end{algorithmic}
\end{algorithm}

\begin{algorithm}
\caption{verifyLanguageOfCategory($treeNode$, $l_X$)}
\begin{algorithmic}
\STATE $r_{curr} \gets treeNode.appliedRule$
\FOR{$c \in treeNode.children$}
	\STATE $currPos \gets treeNode.children.index(c)$
	\IF{$c \in V_{1}\cup\Sigma_{1}$}
		\IF{$currPos \neq pos(h_{X}(r_{curr}), c)$}
			\STATE $EMSentences.delete(l_{X})$
			\RETURN
		\ENDIF
	\ELSIF{$c \in V_{2}\cup\Sigma_{2}$}
		\IF{$currPos \neq pos(h_{12}(h_{X}(r_{curr})), c)$}
			\STATE $EMSentences.delete(l_{X})$
			\RETURN
		\ENDIF
	\ELSE
		\STATE $pos_{1} \gets pos(h_{X}(r_{curr}), f_{X}(c))$
		\STATE $pos_{1} \gets pos(h_{12}(h_{X}(r_{curr})), f_{12}(f_{X}(c)))$
		\IF {$currPos \neq pos_{1}$ and $currPos \neq pos_{2}$}
			\STATE $EMSentences.delete(l_{X})$
			\RETURN
		\ELSIF{$currPos \neq pos_{2}$}
			\STATE $replace(c,f_{X}(c))$
		\ELSE
			\STATE $replace(c,f_{12}(f_{X}(c)))$
		\ENDIF
	\ENDIF
	\STATE $verifyLanguageOfCategory(c)$
\ENDFOR
\RETURN
\end{algorithmic}
\end{algorithm}

\begin{algorithm}
\caption{checkEquivalenceConstraint($treeNode$, $l_X$)}
\begin{algorithmic}
\STATE $r_{curr} \gets treeNode.appliedRule$
\FOR {$i=0$ \TO $treeNode.children.length-2$}
	\STATE $c \gets treeNode.children[i]$
	\STATE $d \gets treeNode.children[i+1]$
	\IF {$(c \in V_{1} \cup \Sigma_{1}$ and $d \in V_{2} \cup \Sigma_{2})$ or \par $(c \in V_{2} \cup \Sigma_{2}$ and $d \in V_{1} \cup \Sigma_{1})$}
		\FOR {$e \in h_{X}(r_{curr}).lhs[0..i]$}
			\IF {$e \in V_1$}
				\IF {$f_{12}(e) \notin h_{12}(h_{X}(r_{curr})).lhs[0..i]$}
					\STATE $EMSentences.delete(l_{X})$
					\RETURN
				\ENDIF
			\ELSE
				\IF {$g_{12}(e) \notin h_{12}(h_{X}(r_{curr})).lhs[0..i]$}
					\STATE $EMSentences.delete(l_{X})$
					\RETURN
				\ENDIF
			\ENDIF
		\ENDFOR
	\ENDIF
\ENDFOR
\FOR {$child \in treeNode.children$}
	\STATE $checkEquivalenceConstraint(child, l_X)$
\ENDFOR	
\RETURN
\end{algorithmic}
\end{algorithm}

\begin{algorithm}
\caption{allowLS($treeNode, allowList$)}
\begin{algorithmic}
\IF{$treeNode \in allowList$}
	\STATE $treeNode.langLabel \gets "either"$
	\RETURN
\ELSE
	\FOR{$c \in treeNode.children$}
		\STATE $allowLS(c, allowList)$
	\ENDFOR
	\IF{$c \in V_{1} \cup \Sigma_{1}$ or $c.langLabel$ is $"either" \forall c \in treeNode.children$}
		\STATE $replace(treeNode,f_{X}(treeNode))$
	\ELSE
    	\STATE ...
	\ENDIF
\ENDIF
\RETURN
\end{algorithmic}
\end{algorithm}

%% file: eacl2017.bbl
\begin{thebibliography}{}

\bibitem[\protect\citename{Adel \bgroup et al.\egroup
  }2013]{adel2013combination}
Heike Adel, Ngoc~Thang Vu, and Tanja Schultz.
\newblock 2013.
\newblock Combination of recurrent neural networks and factored language models
  for code-switching language modeling.
\newblock In {\em ACL (2)}, pages 206--211.

\bibitem[\protect\citename{Adel \bgroup et al.\egroup }2015]{adel2015syntactic}
Heike Adel, Ngoc~Thang Vu, Katrin Kirchhoff, Dominic Telaar, and Tanja Schultz.
\newblock 2015.
\newblock Syntactic and semantic features for code-switching factored language
  models.
\newblock {\em IEEE/ACM Transactions on Audio, Speech, and Language
  Processing}, 23(3):431--440.

\bibitem[\protect\citename{Bali \bgroup et al.\egroup }2014]{bali-2014}
Kalika Bali, Yogarshi Vyas, Jatin Sharma, and Monojit Choudhury.
\newblock 2014.
\newblock {"i am borrowing ya mixing?" an analysis of English-Hindi code mixing
  in Facebook}.
\newblock In {\em {Proc. First Workshop on Computational Approaches to Code
  Switching, EMNLP}}.

\bibitem[\protect\citename{Das and Gamb{\"a}ck}2013]{das2013code}
Amitava Das and Bj{\"o}rn Gamb{\"a}ck.
\newblock 2013.
\newblock Code-mixing in social media text: the last language identification
  frontier.
\newblock {\em Traitement Automatique des Langues (TAL): Special Issue on
  Social Networks and NLP}, 54(3).

\bibitem[\protect\citename{DeNero and Klein}2007]{berkeley-parser-2}
John DeNero and Dan Klein.
\newblock 2007.
\newblock Tailoring word alignments to syntactic machine translation.
\newblock In {\em Proc. ACL}.

\bibitem[\protect\citename{DiSciullo \bgroup et al.\egroup
  }1986]{Discuillo1986}
A.-M. DiSciullo, Pieter Muysken, and R.~Singh.
\newblock 1986.
\newblock Government and code-mixing.
\newblock {\em Journal of Linguistics}, 22:1--24.

\bibitem[\protect\citename{Elfardy \bgroup et al.\egroup
  }2014]{elfardy2014aida}
Heba Elfardy, Mohamed Al-Badrashiny, and Mona Diab.
\newblock 2014.
\newblock Aida: Identifying code switching in informal arabic text.
\newblock {\em EMNLP 2014}, page~94.

\bibitem[\protect\citename{Huang and Yates}2014]{huang2014improving}
Fei Huang and Alexander Yates.
\newblock 2014.
\newblock Improving word alignment using linguistic code switching data.
\newblock In {\em EACL}, pages 1--9.

\bibitem[\protect\citename{Joshi}1985]{Joshi85}
A.~K. Joshi.
\newblock 1985.
\newblock {Processing of Sentences with Intrasentential Code Switching}.
\newblock In D.~R. Dowty, L.~Karttunen, and A.~M. Zwicky, editors, {\em Natural
  Language Parsing: Psychological, Computational, and Theoretical
  Perspectives}, pages 190--205. Cambridge University Press, Cambridge.

\bibitem[\protect\citename{Klein and Manning}2003]{klein-stanford-parser}
Dan Klein and Christopher~D. Manning.
\newblock 2003.
\newblock Accurate unlexicalized parsing.
\newblock In {\em Proceedings of the 41st Meeting of the Association for
  Computational Linguistics}, pages 423--430.

\bibitem[\protect\citename{Liang \bgroup et al.\egroup
  }2006]{berkeley-parser-1}
Percy Liang, Ben Taskar, and Dan Klein.
\newblock 2006.
\newblock Alignment by agreement.
\newblock In {\em Proc. NAACL}.

\bibitem[\protect\citename{Muysken}1995]{Muysken1995}
Pieter Muysken.
\newblock 1995.
\newblock {Code-switching and grammatical theory}.
\newblock In Lesley Milroy and Pieter Muysken, editors, {\em One Speaker, Two
  Languages: Cross-disciplinary Perspectives on Code-switching}, pages
  177--198. Cambridge University Press, Cambridge.

\bibitem[\protect\citename{Myers-Scotton}1993]{Myers-Scotton1993}
Carol Myers-Scotton.
\newblock 1993.
\newblock {\em Duelling Languages:{G}rammatical structure in Code-switching}.
\newblock Clarendon Press, Oxford.

\bibitem[\protect\citename{Myers-Scotton}1995]{Myers-Scotton1995}
Carol Myers-Scotton.
\newblock 1995.
\newblock {A lexically based model of code-switching}.
\newblock In Lesley Milroy and Pieter Muysken, editors, {\em One Speaker, Two
  Languages: Cross-disciplinary Perspectives on Code-switching}, pages
  233--256. Cambridge University Press, Cambridge.

\bibitem[\protect\citename{Poplack}1980]{poplack-1980}
Shana Poplack.
\newblock 1980.
\newblock {Sometimes I’ll start a sentence in Spanish y termino en espanol}.
\newblock {\em Linguistics}, 18:581--618.

\bibitem[\protect\citename{Sankoff}1998]{sankoff1998}
David Sankoff.
\newblock 1998.
\newblock A formal production-based explanation of the facts of code-switching.
\newblock {\em Bilingualism: language and cognition}, 1(01):39--50.

\bibitem[\protect\citename{Sharma \bgroup et al.\egroup
  }2016]{sharma2016shallow}
Arnav Sharma, Sakshi Gupta, Raveesh Motlani, Piyush Bansal, Manish Shrivastava,
  Radhika Mamidi, and Dipti~M. Sharma.
\newblock 2016.
\newblock Shallow parsing pipeline - hindi-english code-mixed social media
  text.
\newblock In {\em Proceedings of the 2016 Conference of the North American
  Chapter of the Association for Computational Linguistics: Human Language
  Technologies}, number~1, pages 1340--1345. Association for Computational
  Linguistics.

\bibitem[\protect\citename{Solorio and Liu}2008]{solorio2008part}
Thamar Solorio and Yang Liu.
\newblock 2008.
\newblock Part-of-speech tagging for english-spanish code-switched text.
\newblock In {\em Proceedings of the Conference on Empirical Methods in Natural
  Language Processing}, pages 1051--1060. Association for Computational
  Linguistics.

\bibitem[\protect\citename{Solorio \bgroup et al.\egroup }2014]{solorio:2014}
Thamar Solorio, Elizabeth Blair, Suraj Maharjan, Steven Bethard, Mona Diab,
  Mahmoud Gohneim, Abdelati Hawwari, Fahad AlGhamdi, Julia Hirschberg, Alison
  Chang, et~al.
\newblock 2014.
\newblock Overview for the first shared task on language identification in
  code-switched data.
\newblock {\em Proceedings of The First Workshop on Computational Approaches to
  Code Switching, EMNLP}, pages 62--72.

\bibitem[\protect\citename{Vyas \bgroup et al.\egroup }2014]{vyas-emnlp-2014}
Yogarshi Vyas, Spandana Gella, Jatin Sharma, Kalika Bali, and Monojit
  Choudhury.
\newblock 2014.
\newblock {POS Tagging of English-Hindi Code-Mixed Social Media Content}.
\newblock In {\em {Proc. EMNLP}}, pages 974--979.

\bibitem[\protect\citename{Y{\i}lmaz \bgroup et al.\egroup
  }2016]{yilmaz2016investigating}
Emre Y{\i}lmaz, Henk van~den Heuvel, and David van Leeuwen.
\newblock 2016.
\newblock Investigating bilingual deep neural networks for automatic
  recognition of code-switching frisian speech.
\newblock {\em Procedia Computer Science}, 81:159--166.

\end{thebibliography}
